\documentclass[11pt,a4paper]{article}
\usepackage[hyperref]{acl2019}
\usepackage{times}
\usepackage{latexsym}
\usepackage{graphicx}
\usepackage[labelformat=simple]{subcaption}
\usepackage{tcolorbox}
\usepackage{todonotes}
\usepackage{xcolor}
\usepackage{amsmath}
\usepackage{url}
\usepackage[export]{adjustbox}[2011/08/13]

\DeclareMathOperator*{\argmax}{arg\,max}

\aclfinalcopy 

\title{Pre-Learning Environment Representations for Data-Efficient Neural Instruction Following}

\author{David Gaddy \and Dan Klein \\
  Computer Science Division \\
  University of California, Berkeley \\
  {\tt \{dgaddy,klein\}@berkeley.edu} \\}

\date{}

\begin{document}
\maketitle
\begin{abstract}

We consider the problem of learning to map from natural language instructions to state transitions (actions) in a data-efficient manner.  Our method takes inspiration from the idea that it should be easier to ground language to concepts that have already been formed through pre-linguistic observation.  We augment a baseline instruction-following learner with an initial environment-learning phase that uses observations of language-free state transitions to induce a suitable latent representation of actions before processing the instruction-following training data.  We show that mapping to pre-learned representations substantially improves performance over systems whose representations are learned from limited instructional data alone.

\end{abstract}

\section{Introduction}

In the past several years, neural approaches have become increasingly central to the instruction following literature \cite[e.g.][]{misra2018mapping, chaplot2018gated, mei2016listen}.
However, neural networks' powerful abilities to induce complex representations have come at the cost of data efficiency.
Indeed, compared to earlier logical form-based methods, neural networks can sometimes require orders of magnitude more data.
The data-hungriness of neural approaches is not surprising -- starting with classic logical forms improves data efficiency by presenting a system with pre-made abstractions, where
end-to-end neural approaches must do the hard work of inducing abstractions on their own.
In this paper, we aim to combine the power of neural networks with the data-efficiency of logical forms by pre-learning abstractions in a semi-supervised way, satiating part of the network's data hunger on cheaper unlabeled data from the environment.

\begin{figure}
    \centering
    \includegraphics[scale=.16]{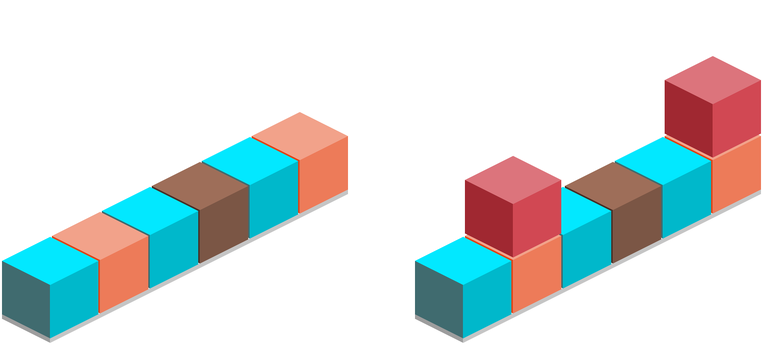}
    \caption{After seeing this transition, a neural net might generalize this action as \emph{stack red blocks to the right of blue blocks except for on brown blocks}, but a generalization like \emph{stack red blocks on orange blocks} is more plausible and generally applicable.  We aim to guide our model towards more plausible generalizations by pre-learning inductive biases from observations of the environment.}
    \label{fig:generalization_example}
\end{figure}

When neural nets have only limited data that pairs language with actions, they suffer from a lack of inductive bias,
 fitting the training data but generalizing in ways that seem nonsensical to humans.
For example, a neural network given the transition shown in Figure~\ref{fig:generalization_example} might map the corresponding instruction to an adequate but unlikely meaning that red blocks should be stacked to the right of blue blocks except for on brown blocks.
The inspiration for this work comes from the idea that humans avoid spurious hypotheses like this example partly because they have already formed a set of useful concepts about their environment before learning language \cite{bloom2000children,hespos2004conceptual}.
These pre-linguistic abstractions then constrain language learning and help generalization.

\begin{figure*}
    \centering
    \begin{subfigure}[b]{.49\textwidth}
    \centering
    \includegraphics[width=.98\textwidth]{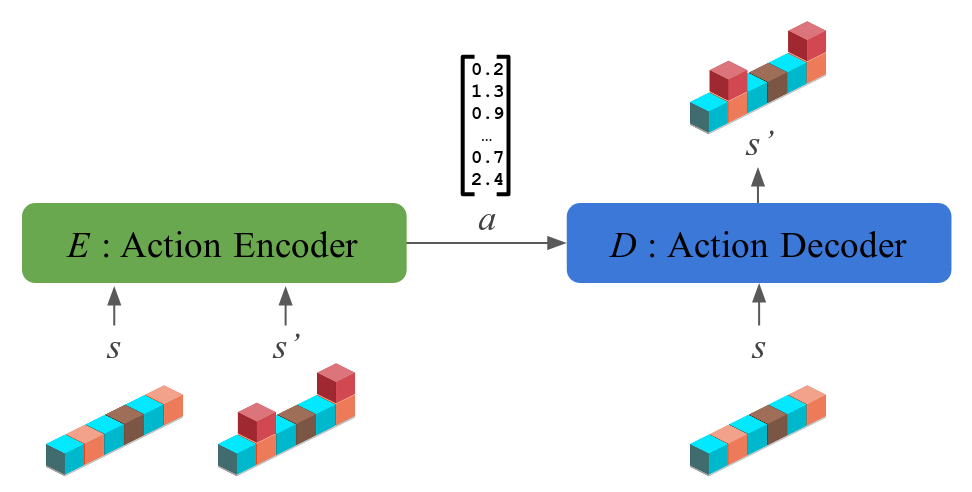}
    \vspace{.05cm}
    \caption{Environment learning}
    \label{fig:pretraining}
    \end{subfigure}
    \begin{subfigure}[b]{.49\textwidth}
    \centering
    \includegraphics[width=\textwidth]{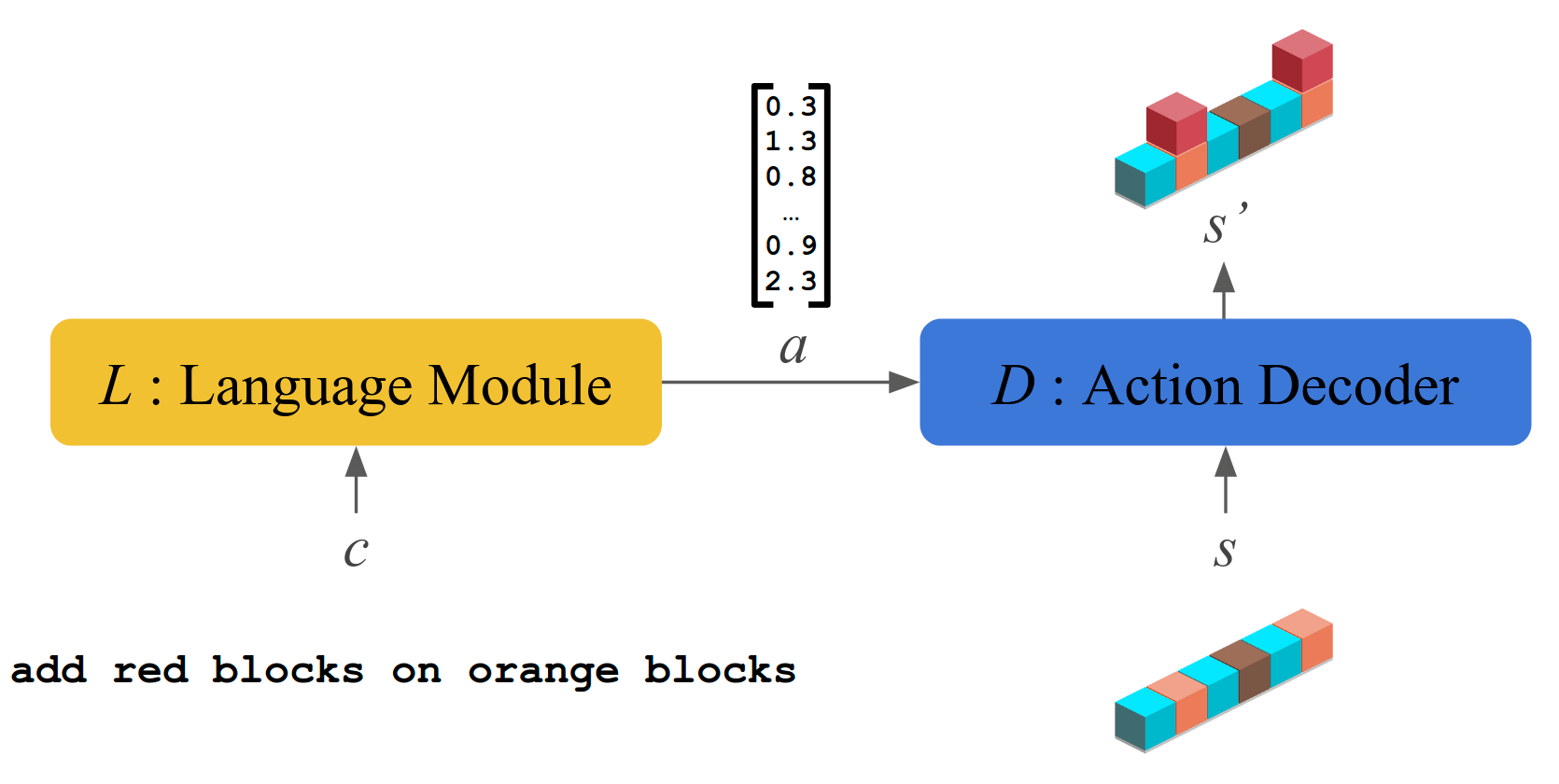}
    \caption{Language learning}
    \label{fig:languagelearning}
    \end{subfigure}
    \caption{Diagram of the network modules during the environment learning and language learning phases.  $s$ and $s'$ represent states before and after an action, $c$ represents a natural language command, and $a$ represents a latent action representation.  The environment learning phase \subref{fig:pretraining} uses a conditional autoencoder to pre-train the decoder $D$ toward a good representation space for $a$, so that fewer linguistic examples are needed during language learning \subref{fig:languagelearning}.}
    \label{fig:overview}
\end{figure*}

With this view in mind, we allow our instruction following agent to observe the environment and build a representation of it prior to seeing any linguistic instructions.
In particular, we adopt a semi-supervised setup with two phases, as shown in Figure~\ref{fig:overview}: an \emph{environment learning} phase where the system sees samples of language-free state transitions from actions in the environment, and a \emph{language learning} phase where instructions are given along with their corresponding effects on the environment.
This setup applies when interactions with the environment are plentiful but only a few are labeled with language commands.
For example, a robotic agent could passively observe a human performing a task, without requiring the human to perform any work they would not normally do, so that later the agent would need less direct instruction from a human in the form of language.
We present an environment learning method that uses observations of state transitions to build a representation that aligns well with the transitions that tend to occur.
The method takes advantage of the fact that in complex environments (or even relatively simple ones), not every state transition is equally likely, but
the patterns of actions that do occur
hint at an underlying structure that we can try to capture.

We demonstrate the effectiveness of our pre-trained representations by using them to increase data efficiency on two instruction-following tasks (Section~\ref{sec:experiments}).
We show that when given few instruction examples, a network using our pre-learned representations performs substantially better than an otherwise identical network without these representations, increasing performance by over ten absolute percentage points on small datasets and increasing data-efficiency by more than an order of magnitude.
We find that while performance with a typical neural representation trained end-to-end lags considerably behind performance with human-designed representations, our unsupervised representations are able to help cross a substantial portion of this gap.
In addition, we perform analysis of the meaning captured by our representations during the unsupervised environment learning phase, demonstrating that the semantics captured has noteworthy similarity to a hand-defined system of logical forms (Section~\ref{sec:generalization}).

\section{Problem Setup}

This work applies to the class of problems where instructions are mapped to actions conditioned on an environment state.
These tasks can be formalized as learning a mapping $M(s,c)\mapsto s'$, where $c$ is a command in natural language, $s$ is an environment state, and $s'$ is the desired environment state after following the command.
Classically, these problems are approached by introducing a logical form $l$ that does not depend on the state $s$, and learning a mapping from language to logical forms $P(c)\mapsto l$ \cite{artzi2013weakly,zettlemoyer2005learning}.
A hand-defined execution function $Q(s,l)\mapsto s'$ is then used to generalize the action across all possible states.
When $P$ and $Q$ are composed, $Q$ constrains the overall function to generalize in a semantically coherent way across different states.

In contrast to the logical form-based method, our work builds off of an end-to-end neural approach, which is applicable in settings where a system of logical forms is not provided.
We structure our network as a language module $L$ and action decoder $D$ in an encoder-decoder style architecture (Figure~\ref{fig:languagelearning}), similar to previous neural instruction following work \cite[e.g.][]{mei2016listen}.
$L$ and $D$ are analogous to the $P$ and $Q$ functions in the logical form approach, however, unlike before, the interface between the two modules is a vector $a$ and the function $D$ is learned.
This gives the neural network greater flexibility, but also creates the problem that the decoder $D$ is no longer constrained to generalize across different states in natural ways.

\section{Method}

\subsection{Learning Action Representations from the Environment}
\label{sec:env-learning}

The goal of this paper is improve data efficiency by pre-training the decoder $D$ to use a better-generalizing representation for the vector $a$.  We do this in an unsupervised way by allowing our system to see examples of state transitions (actions) in the environment before seeing any language.
We suppose the existence of a large number of language-free state transitions $s,s'$ and introduce an environment learning phase to learn representations of these transitions before language learning starts.
During this environment learning phase, we train a \emph{conditional autoencoder} of $s'$ given $s$ by introducing an additional encoder $E(s,s')\mapsto a$ to go along with decoder $D(s,a)\mapsto s'$, as shown in Figure~\ref{fig:pretraining}.
Both $E$ and $D$ are given the initial state $s$, and $E$ must create a representation of the final state $s'$ so that $D$ can reproduce it from $s$.
The parameters of $E$ and $D$ are trained to maximize log likelihood of $s'$ under the output distribution of $D$.
\begin{equation}
\argmax_{\theta_E,\theta_D}\left[\log P_D(s'|s,E(s,s')\right]
\end{equation}
If given enough capacity, the representation $a$ might encode all the information necessary to produce $s'$, allowing the decoder to ignore $s$.
However, with a limited representation space, the decoder must learn to integrate information from $a$ and $s$, leading $a$ to capture an abstract representation of the transformation between $s$ and $s'$.
To be effective, the representation $a$ needs to be widely applicable in the environment and align well with the types of state transitions that typically occur.
These pressures cause the representation to avoid meanings like \emph{to the right of blue except for on brown} that rarely apply.
Note that during pre-training, we do not add any extra information to indicate that different transitions might be best represented with the same abstract action, but the procedure described here ends up discovering this structure on its own.

Later, after demonstrating the effectiveness of this environment learning procedure in Section~\ref{sec:experiments}, we introduce two additional improvements to the procedure in sections \ref{sec:discrete} and \ref{sec:encoder-matching}.
In Section \ref{sec:generalization}, we show that our pre-training discovers representations that align well with logical forms when they are provided.

\subsection{Language Learning}
\label{sec:lang-learning}

After environment learning pre-training, we move to the language learning phase.  In the language learning phase, we are given state transitions paired with commands ($s,s',c$) and learn to map language to the appropriate result state $s'$ for a given state $s$.
As discussed above and shown in Figure~\ref{fig:languagelearning}, we form an encoder-decoder using a language encoder $L$ and action decoder $D$.
To improve generalization, we use the decoder $D$ that was pre-trained during environment learning.
If $D$ generalizes representations across different states in a coherent way as we hope, then the composed function $D(s,L(c))$ will also generalize well.
We can either fix the parameters of $D$ after environment learning or simply use the pre-learned parameters as initialization, which will be discussed more in the experiments section below.
The language module $L$ is trained by differentiating through the decoder $D$ to maximize the log probability that $D$ outputs the correct state $s'$.

\begin{equation}
\label{eqn:languageloss}
\argmax_{\theta_L}\left[\log P_D(s'|s,L(c))\right]
\end{equation}

\subsection{Comparison with Action Priors}
One of the roles of environment learning pre-training is to learn something like a prior over state transitions, ensuring that we select a reasonable action based on the types of transitions that we have seen.
However, the method described here has advantages over a method
that just learns a transition prior.
In addition to representing which transitions are likely, our pre-training method also induces structure within the space of transitions.
A single action representation $a$ can be applied to many different states to create different transitions, effectively creating a group of transitions.  After training, this grouping might come to represent a semantically coherent category (see analysis in Section~\ref{sec:generalization}).
This type of grouping information may not be easily extractable from a prior.
For example, a prior can tell you that stacking red blocks on orange blocks is likely across a range of initial configurations, but our pre-training method may also choose to represent all of these transitions with the same vector $a$.
Finding this underlying structure is key to the generalization improvements seen with our procedure.

\section{Experiments}
\label{sec:experiments}

We evaluate our method in two different environments, as described below in sections \ref{sec:stacks} and \ref{sec:regex}.\footnote{Code for all experiments can be found at \texttt{github.com/dgaddy/environment-learning}.}

\subsection{Block Stacking}
\label{sec:stacks}
For our first test environment, we use the block stacking task introduced by \citet{wang2016learning} and depicted in Figure~\ref{fig:generalization_example}.
This environment consists of a series of levels (tasks), where each level requires adding or removing blocks to get from a start configuration to a goal configuration.
Human annotators were told to give the computer step by step instructions on how to move blocks from one configuration to the other.
After each instruction, the annotator selected the desired resulting state from a list.

Following the original work for this dataset \cite{wang2016learning}, we adopt an online learning setup and metric.
The data is broken up into a number of sessions, one for each human annotator, where each session contains a stream of commands $c$ paired with block configuration states $s$.
The stream is processed sequentially, and
for each instruction the system predicts the result of applying command $c$ to state $s$, based on a model learned from previous examples in the stream.
After making a prediction, the system is shown the correct result $s'$ and is allowed to make updates to its model before moving on to the next item in the stream.
The evaluation metric, online accuracy, is then the percentage of examples for which the network predicted the correct resulting state $s'$ when given only previous items in the stream as training.
Under this metric, getting predictions correct at the beginning of the stream, when given few to no examples, is just as important as getting predictions correct with the full set of data, making it as much a measure of data-efficiency as of final accuracy.
The longest sessions only contain on the order of 100 training examples, so the bulk of predictions are made with \emph{only tens of examples}.

To train a neural model in this framework, the model is updated by remembering all previous examples seen in the stream so far and training the neural network to convergence on the full set of prior examples.
While training the network to convergence after every example is not very computationally efficient, the question of making efficient online updates to neural networks is orthogonal to the current work, and we wish to avoid any confounds introduced by methods that make fewer network updates.

Since the original dataset does not contain a large number of language-free state transitions as we need for environment learning, we generate synthetic transitions.
To generate state transitions $s,s'$, we generate new levels using the random procedure used in the original work and programmatically determine a sequence of actions that solve them.
The levels of the game are generated by a procedure which selects random states and then samples a series of transformations to apply to generate a goal state.
We create a function that generates a sequence of states from the start to the goal state based on the transformations used during goal generation.
Most of the levels require one or two actions with simple descriptions to reach the goal.
Following the assumption that state transitions in the environment are plentiful, we generate new transitions for every batch during environment learning.  We leave an analysis of the effect of environment learning data size to future work.

\subsubsection{State Representation and Network Architecture}
\label{sec:stacks-arch}

We represent a state as a two dimensional grid, where each grid cell represents a possible location (stack index and height) of a block.  The state inputs to the encoder and decoder networks use a one-hot encoding of the block color in each cell or an empty cell indicator if no block is present.  The output of the decoder module is over the same grid, and a softmax over colors (or empty) is used to select the block at each position.
Note that the original work in this environment restricted outputs to states reachable from the initial state by a logical form, but here we allow any arbitrary state to be output and the model must learn to select from a much larger hypothesis space.

The encoder module $E$ consists of convolutions over the states $s$ and $s'$, subtraction of the two representations, pooling over locations, and finally a fully connected network which outputs the representation $a$.
The decoder module $D$ consists of convolution layers where the input is state $s$ and where $a$ is broadcast across all positions to an intermediate layer.
The language module $L$ runs an LSTM over the words, then uses a fully connected network to convert the final state to the representation $a$.
Details of the architecture and hyperparameters can be found in Appendix~\ref{appendix:block-arch}.

\subsubsection{Results}

Our primary comparison is between a neural network with pre-trained action representations and an otherwise identical neural model with no pre-trained representations.
The neural modules are identical, but in the full model we have fixed the parameters of the decoder $D$ after learning good representations with the environment learning procedure.
We tune the baseline representation size independently since it may perform best under different conditions, choosing among a large range of comparable sizes (details in Appendix~\ref{appendix:block-arch}).
To evaluate the quality of our representations, we also compare with a system using hand-designed logical representations \cite{wang2016learning}.
While not strictly an upper bound, the human-designed representations were designed with intimate knowledge of the data environment and so provide a very good representation of actions people might take.
This makes them a strong point of comparison for our unsupervised action representations.

Table~\ref{tab:results-stacks} shows the results on this task.  We find that training the action representation with environment learning provides a very large gain in performance over an identical network with no pre-linguistic training, from 17.9\% to 25.9\%.
In sections \ref{sec:discrete} and \ref{sec:encoder-matching} below, we'll add discrete representations and an additional loss term which together bring the accuracy to 28.5\%, an absolute increase of more than 10\% over the baseline.
Comparing against the system with human-designed representations shows that the environment learning pre-training substantially narrows the performance gap between hand designed representations and representations learned as part of an end-to-end neural system.

\begin{table}
\centering
\begin{tabular*}{\columnwidth}{l@{\extracolsep{\fill}}r}
 \hline
\multicolumn{2}{c}{Learned Representations (this work)} \\
Baseline & 17.9 \\
Environment Learning & 25.9 \\
\enspace + Discrete $a$ (Section \ref{sec:discrete}) & 27.6 \\
\enspace + Encoder matching (Section \ref{sec:encoder-matching}) & \textbf{28.5} \\ \hline
\multicolumn{2}{c}{Human-Designed Representations} \\
\citet{wang2016learning} & 33.8 \\ \hline
\end{tabular*}

\caption{Online accuracy for the block stacking task.\footnotemark  \phantom{0}  Pre-learning action representations with environment learning greatly improves performance over the baseline model, substantially narrowing the gap between hand designed representations and representations learned as part of an end-to-end neural system.
Note that these numbers represent accuracy after learning from only tens of examples.}
\label{tab:results-stacks}
\end{table}

\subsection{String Manipulation}
\label{sec:regex}
The second task we use to test our method is string manipulation.
In this task a state $s$ is a string of characters and actions correspond to applying a transformation that inserts or replaces characters in the string, as demonstrated in Figure~\ref{fig:regex-example}.
We use the human annotations gathered by \citet{andreas2018learning}, but adapt the setup to better measure data-efficiency.

The baseline neural model was unable to learn useful models for this task using data sizes appropriate for the online learning setup we used in the previous task, so we instead adopt a slightly different evaluation where accuracy at different data sizes is compared.  We structure the data for evaluation as follows:
First, we group the data so that each group contains only a small number of instructions (10).  In the original data, each instruction comes with multiple example strings, so we create distinct datapoints $s,s',c$ for each example with the instruction string repeated.  Our goal is to see how many examples are needed for a model to learn to apply a set of 10 instructions.
We train a model on training sets of different sizes and evaluate accuracy on a held-out set of 200 examples.
We are primarily interested in generalization across new environment states, so the held-out set consists of examples with the same instructions but new initial states $s$.
Due to high data requirements of the baseline neural system, we found it necessary to augment the set of examples for each instruction with additional generated examples according to the regular expressions included with the dataset.
Our final metric is the average accuracy across 5 instruction groups, and we plot this accuracy for different training set sizes.

\begin{figure}
    \centering
    \small
    \begin{tcolorbox}[left=0mm,right=0mm,top=0mm,bottom=0mm]
    \begin{tabular}{ll}
    $c$ & replace consonants with p x \\
    $s$ & \texttt{fines}\\
    $s'$ & \texttt{pxipxepx}\\
    \end{tabular}
    \end{tcolorbox}
    \begin{tcolorbox}[left=0mm,right=0mm,top=0mm,bottom=0mm]
    \begin{tabular}{ll}
    $c$ & add a letter k before every b\\
    $s$ & \texttt{rabbles}\\
    $s'$ & \texttt{rakbkbles}\\
    \end{tabular}
    \end{tcolorbox}
    \begin{tcolorbox}[left=0mm,right=0mm,top=0mm,bottom=0mm]
    \begin{tabular}{ll}
    $c$ & replace vowel consonant pairing with v g\\
    $s$ & \texttt{thatched}\\
    $s'$ & \texttt{thvgchvg}\\
    \end{tabular}
    \end{tcolorbox}
    \begin{tcolorbox}[left=0mm,right=0mm,top=0mm,bottom=0mm]
    \begin{tabular}{ll}
    $c$ & add b for the third letter\\
    $s$ & \texttt{thanks}\\
    $s'$ & \texttt{thbanks}\\
    \end{tabular}
    \end{tcolorbox}
    
    \caption{Examples from the string manipulation task along with desired outputs.}
    \label{fig:regex-example}
\end{figure}

\footnotetext{Although the variance between runs was small relative to the gaps in performance, we report an average over three random initializations to ensure a fair comparison.}

State transitions for environment learning are generated synthetically by selecting words from a dictionary and applying regular expressions, where the regular expressions to apply were sampled from a regular-expression generation procedure written by the creators of the original dataset.
The environment learning procedure is exposed to transitions from thousands of unique regular expressions that it must make sense of and learn to represent.

\subsubsection{Network Architecture}

For this task, the state inputs and outputs are represented as sequences of characters.
The encoder $E$ runs a LSTM over the character sequences for $s$ and $s'$, then combines the final states with a feedforward network to get $a$.  The decoder $D$ runs a LSTM over the characters of $s$, combines this with the representation $a$, then outputs $s'$ using another LSTM.
The module architecture details and hyperparameters can be found in Appendix~\ref{appendix:string-arch}.

Since our evaluation for this task considers larger dataset sizes in addition to very small sizes, we do not fix the parameters of the decoder $D$ as we did in the previous task, but instead use the pre-trained decoder as initialization and train it along with the language module parameters.  Allowing the parameters to train gives the decoder more power to change its representations when it has enough data to do so, while the initialization helps it generalize much better, as demonstrated by our results below.

\subsubsection{Results}

As with the other dataset (Section~\ref{sec:stacks}), we compare the full model with a baseline that has no environment learning, but an otherwise identical architecture.
To ensure a fair comparison, we tune the baseline representation size separately, choosing the best from a range of comparable sizes (see Appendix~\ref{appendix:string-arch}).

Figure~\ref{fig:regex-results} plots the accuracy across different data sizes of the baseline neural model and the model with environment learning pre-training.
Note that models are trained to convergence, so this plot is intended to indicate data efficiency, not training speed (though training speed is also likely to increase at similar rates).
As seen in the figure, environment learning substantially increases data efficiency on this task.
At small data sizes, the baseline model struggles to generalize across different states $s$, often choosing to output one of the training outputs $s'$ rather than learning a rule and applying it to $s$.
Environment learning greatly increases the ability of the model to find the correct generalization.

\begin{figure}
    \centering
    \includegraphics[width=\columnwidth]{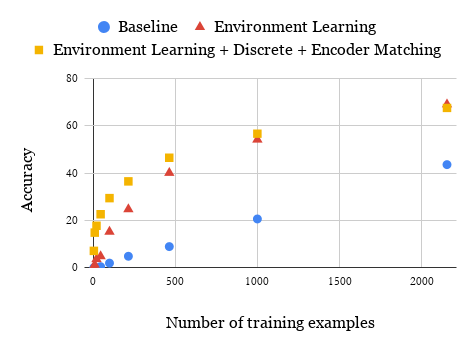}
    \caption{Accuracy for the string manipulation task as the number of examples ($s,s',c$) is increased.  Environment learning pre-training increases data efficiency by an order of magnitude or more.  The results in yellow include additional improvements described in sections \ref{sec:discrete} and \ref{sec:encoder-matching} below.}
    \label{fig:regex-results}
\end{figure}

\section{Discrete Action Representations}
\label{sec:discrete}

In this section, we describe a variant of our model where we use a discrete representation $a$ instead of a continuous one and evaluate this variant on our two tasks.
Semantics is often defined in terms of discrete logical structures.
Even in continuous environments, it is often natural to describe objects and relations in discrete ways.
Using a discrete space for our learned action representations can provide useful inductive bias for capturing this discrete structure.
In addition, a discrete representation has the potential advantages of increased robustness and increased control of information flow during environment learning.

When using discrete representations, we divide our $a$ into $n$ different discrete random variables where each variable selects its value from one of $k$ categories.
We train the discrete representation using the Gumbel-Softmax
\cite{jang2017categorical,maddison2017concrete}, which gives us a continuous relaxation of the discrete variables that we can backpropagate through.
The Gumbel-Softmax operation transforms an $n\times k$ vector into $n$ discrete random variables, which we represent as one-hot vectors and feed through the rest of the network just as we would a continuous representation.
The Gumbel-Softmax is calculated as
$$G(x_i)=\frac{\exp(x_i+\epsilon_i)}{\sum_{j=0}^k\exp(x_j+\epsilon_j)}$$
where $\epsilon$ are i.i.d.\ samples from the Gumbel(0,1) distribution and the vector $x$ represents un-normalized log probabilities for each of the $k$ categories.
This operation is analogous to a softening of a sample from the distribution.
While the original work suggested the use of an annealed temperature parameter, we did not find it necessary in our experiments.
We use the straight-through variant, where the discrete mode of the softmax distribution is used in the forward pass, but the backward pass is run as if we had used the continuous value $G(x_i)$.  We found that a representation with $n=20$ variables and $k=30$ values works well for all our experiments.

Using discrete representations instead of continuous representations further improves environment learning results on both tasks, increasing the block stacking task accuracy from 25.9\% to 27.6\% (Table~\ref{tab:results-stacks}) and improving string manipulation on moderate training sizes (200 examples) from 24.7\% to 36.9\%.
We also ran the baseline neural models with discrete representations for comparison but did not observe any performance gains, indicating that the discrete representations are useful primarily when used with environment learning pre-training.

\section{Encoder Representation Matching}
\label{sec:encoder-matching}

One potential difficulty that may occur when moving from the environment learning to the language learning phase is that the language module $L$ could choose to use parts of the action representation space that were not used by the encoder during environment learning.
Because the decoder has not seen these representations, it may not have useful meanings associated with them, causing it to generalize in a suboptimal way.
In this section, we introduce a technique to alleviate this problem and show that it can lead to an additional improvement in performance.

Our fix uses an additional loss term to encourage the language module $L$ to output representations that are similar to those used by the encoder $E$.
For a particular input $c,s,s'$ in the language learning phase, we run the encoder on $s,s'$ to generate a possible representation $a_E$ of this transition.
We then add an objective term for the
log likelihood of $a_E$ under $L$'s output distribution.  The full objective during language learning is then
\begin{equation}
\label{eqn:full-loss}
\argmax_{\theta_L}\left[\log P_D(s'|s,L(c))+\lambda \log P_L(a_E|c)\right]
\end{equation}
where the encoder matching weight $\lambda$ is a tuned constant.
$P_L$ is the softmax probability from the output of the language module when using discrete representations for $a$, and $a_E$ is the discrete mode of the encoder output distribution.\footnote{When using continuous representations, a $\ell_2$ distance penalty could be used to encourage similarity between the output of $L$ and $E$, though this tended to be less effective in our experiments.}

Using this technique on the block stacking task (with $\lambda=.01$), we see a performance gain of .9\% over discrete-representation environment learning to reach an accuracy of 28.5\%.  This number represents our full model performance and demonstrates more than 10\% absolute improvement over the baseline.  The additional loss also provides gains on string manipulation, especially on very small data sizes (e.g. from 3.9\% to 14.8\% with only 10 examples).  The performance curve of our complete model is shown in Figure~\ref{fig:regex-results}.  With our full model, it takes less than 50 examples to reach the same performance as with 1000 examples using a standard neural approach.

\section{Exploring the Learned Representation}
\label{sec:generalization}

A primary goal of the environment learning procedure is to find a representation of actions that generalizes in a semantically minimal and coherent way.
In this section, we perform analysis to see what meanings the learned action representations capture in the block stacking environment.
Since logical forms are engineered to capture semantics that we as humans consider natural, we compare our learned representations with a system of logical forms to see if they capture similar meanings without having been manually constrained to do so.
We compare the semantics of the learned and logical representations by comparing their effect on different states, based on the method of \citet{andreas2017analogs}.

We test an encoder and decoder using the following procedure:
First, we generate a random transition $s_1,s_1'$ from the same distribution used for environment learning and run the encoder to generate an action representation $a_1$ for this transition.
Then, we generate a new state $s_2$ from the environment and run the decoder on the new state with the representation generated for the original state: $D(a_1,s_2)\mapsto \bar{s}_2$.
We are interested in whether the output $\bar{s}_2$ of this decoding operation corresponds to a generalization that would be made by a simple logical form.
Using a set of logical forms that correspond to common actions in the block stacking environment, we find all simple logical forms that apply to the original transition $s_1,s_1'$ and all forms that apply to the predicted transition $s_2,\bar{s}_2$.
If the intersection of these two sets of logical forms is non-empty, then the decoder's interpretation of the representation $a_1$ is consistent with some simple logical form.
We repeat this procedure on 10,000 state transitions to form a logical form consistency metric.

Running this test on our best-performance model, we find that 84\% of the generalizations are consistent with one of the simple logical forms we defined.  This result indicates that while the generalization doesn't perfectly match our logical form system, it does have a noteworthy similarity.
An inspection of the cases that did not align with the logical forms found that the majority of the ``errors'' could in fact be represented by logical forms, but ones that
 were not minimal.
In these cases, the generalization isn't unreasonable, but has slightly more complexity than is necessary.
For example, from a transition that could be described either as \emph{stack a blue block on the leftmost block} or separately as \emph{stack blue blocks on red blocks} (where red only appears in the leftmost position), the representation $a$ that is generated generalizes across different states as the conjunction of these two meanings (\emph{stack blue blocks on the leftmost block AND on red blocks}), even though no transitions observed during environment learning would need this extra complexity to be accurately described.

\section{Related Work}

Many other works use autoencoders to form representations in an unsupervised or semi-supervised way.
Variants such as denoising autoencoders \cite{vincent2008extracting} and variational autoencoders \cite{kingma2013auto} have been used for various vision and language tasks.
In the area of semantic grounding, \citet{kocisky2016semantic} perform semi-supervised semantic parsing using an autoencoder where the latent state takes the form of language.

Our approach also relates to recent work on learning artificial languages by simulating agents interacting in an environment \cite[i.a.]{mordatch2018emergence,das2017learning,kottur2017natural}.
Our environment learning procedure could be viewed as a language learning game where the encoder is a speaker and the decoder is a listener.  The speaker must create a ``language'' $a$ that allows the decoder to complete a task.
Many of these papers have found that it is possible to induce representations that align semantically with language humans use, as explored in detail in  \citet{andreas2017analogs}.  Our analysis in Section~\ref{sec:generalization} is based on the method from this work.

Model-based reinforcement learning is another area of work 
that improves data-efficiency by learning from observations of an environment \cite{wang2018look, deisenroth2013survey, kaiser2019model}.
It differs from the current work in which aspect of the environment it seeks to capture: in model-based RL the goal is to model which states will result from taking a particular action, but in this work we aim to learn patterns in what actions tend to be chosen by a knowledgeable actor.

Another related line of research uses language to guide learning about an environment \cite{branavan2012learning,srivastava2017joint,andreas2018learning,hancock2018training}.  These papers use language to learn about an environment more efficiently, which can be seen as a kind of inverse to our work, where we use environment knowledge to learn language more efficiently.

Finally, recent work by \citet{leonandya2018fast} also explores neural architectures for the block stacking task we used in section \ref{sec:stacks}.  The authors recognize the need for additional inductive bias, and introduce this bias by creating additional synthetic \emph{supervised} data with artificial language, creating a transfer learning-style setup.  This is in contrast to our  unsupervised pre-training method that does not need language for the additional data.  Even with their stronger data assumptions, their online accuracy evaluation reaches just 23\%, compared to our result of 28.5\%, providing independent verification of the difficulty of this task for neural networks.

\section{Conclusion}

It is well known that neural methods do best when given extremely large amounts of data.
As a result, much of AI and NLP community has focused on making larger and larger datasets, but we believe it is equally important to go the other direction and explore methods that help performance with little data.
This work introduces one such method.  Inspired by the idea that it is easier to map language to pre-linguistic concepts, we show that when grounding language to actions in an environment, pre-learning representations of actions can help us learn language from fewer language-action pairings.

\section*{Acknowledgments}

This work is supported by the DARPA Explainable Artificial Intelligence (XAI) program.  We would like to thank the members of the Berkeley NLP group and the anonymous reviewers for their helpful feedback.

\bibliography{references}
\bibliographystyle{acl_natbib}

\appendix

\section{Neural Architectures and Hyperparameters}

\subsection{Block Stacking}
\label{appendix:block-arch}

\begin{figure}[b]
    \centering
    \begin{subfigure}[b]{\columnwidth}
    \centering
    \includegraphics[scale=.18]{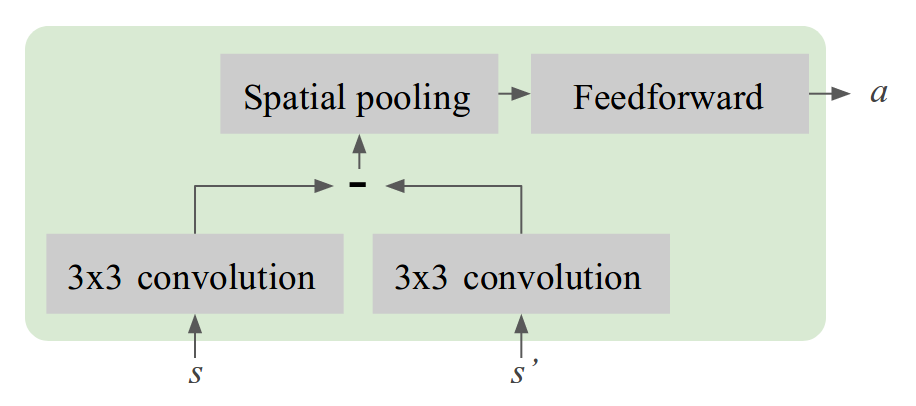}
    \caption{Encoder $E$}
    \end{subfigure}
    \begin{subfigure}[b]{\columnwidth}
    \centering
    \includegraphics[scale=.18]{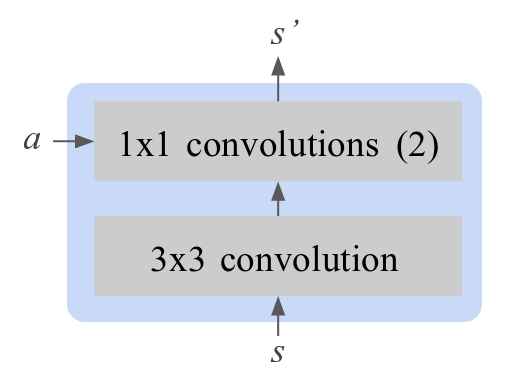}
    \caption{Decoder $D$}
    \end{subfigure}
    \caption{Architecture for block stacking task modules.}
    \label{fig:stacks-arch}
\end{figure}

The encoder and decoder module architectures for the block stacking task are shown in Figure~\ref{fig:stacks-arch}.
The encoder module $E$ consists of convolutions over the states $s$ and $s'$, subtraction of the two representations, pooling over locations, and finally a fully connected network which outputs the representation $a$.
The fully connected network has a single hidden layer.
The decoder module $D$ consists of two convolution layers where the input is state $s$ and where $a$ is broadcast across all positions and concatenated with the input to the second layer.
All convolutions and feedforward layers for $E$ and $D$ have dimension 200 and all intermediate layers are followed by ReLU non-linearities.  Dropout with probability 0.5 was used on the encoder feedforward hidden layer and before the last convolution layer in the decoder.

\begin{figure}
  \centering
  \includegraphics[scale=.15]{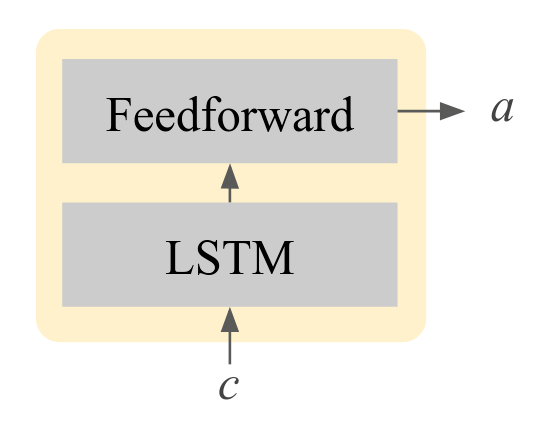}
  \caption{The language module $L$ used for both the block stacking and string manipulation tasks uses a LSTM over the words of the command $c$.}
  \label{fig:language-module}
\end{figure}

The language module $L$ uses a LSTM encoder (Figure~\ref{fig:language-module}).
It takes a command $c$ as a sequence of learned word embeddings, runs an LSTM over them, then projects from the final cell state to get the output vector $a$.
The word embeddings have dimension 100 and the LSTM has hidden size 200.

When using a continuous action representation, $a$ has dimension 600.
When using a discrete representation, we use $n=20$ discrete variables where each takes one of $k=30$ values.
Environment learning is run on 500,000 batches of size 20, after which we fix the parameters of $D$.  During language learning, we optimize $L$ for 50 epochs after each new example is presented, using a batch size of 1.
All optimization is done using Adam with learning rate $0.001$.

To ensure a fair comparison with the baseline, we ran the baseline system with both continuous and discrete representations and took the best.  Generally, the baseline performed slightly better with continuous representations.  We ran with continuous sizes 20, 50, 100, 300, and 600; selecting the best result.  This range was chosen to be between the number of discrete variables $n$ and the total number of inputs to the discretization $n\times k$.

\subsection{String Manipulation}
\label{appendix:string-arch}

\begin{figure}
    \centering
    \begin{subfigure}[b]{.9\columnwidth}
    \centering
    \includegraphics[scale=.18]{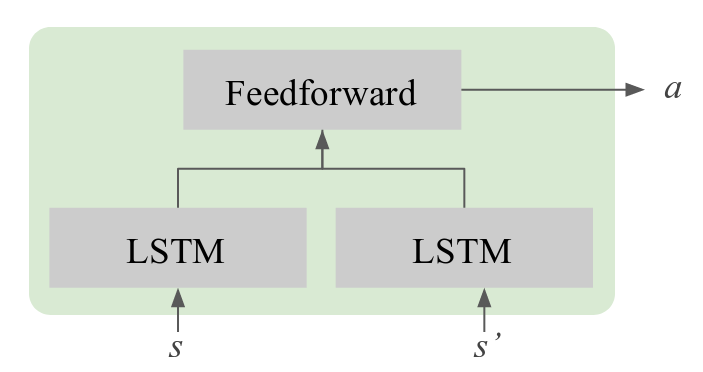}
    \caption{Encoder $E$}
    \end{subfigure}
    \begin{subfigure}[b]{.9\columnwidth}
    \centering
    \includegraphics[scale=.18]{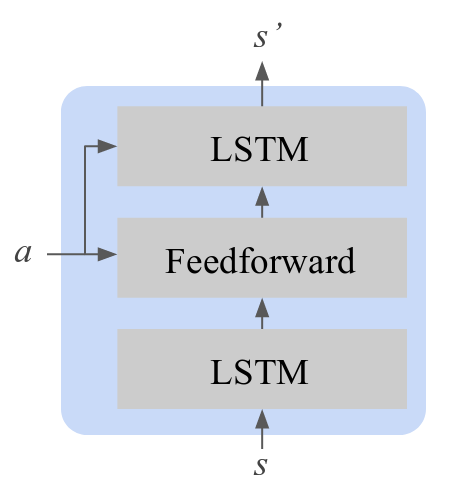}
    \caption{Decoder $D$}
    \end{subfigure}
    \caption{Architecture for string manipulation task modules.}
    \label{fig:regex-arch}
\end{figure}

Figure~\ref{fig:regex-arch} shows the encoder and decoder module architectures for the string manipulation task.
The encoder $E$ runs a LSTM over the character sequences for $s$ and $s'$, using separate LSTMs for the two sequences, but tying their parameters.  The final states of the two LSTMs are then concatenated and fed into a feedforward network with one hidden layer that outputs the action representation $a$.
The decoder $D$ consists of a LSTM over the sequence $s$, a feedforward network of a single linear layer combining $a$ with the LSTM final state, and a LSTM that outputs the sequence $s'$, where the output LSTM's initial state comes from the output of the feedforward network.
$a$ is also concatenated with the previous output embedding that is fed into the input of the LSTM at each timestep.
The character embeddings input to the LSTM have dimension 50, and all LSTM and feedforward layers have dimension 500.
When using a continuous representation $a$, we use a representation dimension of 20, though the results were not overly sensitive to this value.
When using a discrete representation, we use $n=20$ variables where each takes one of $k=30$ values.

The language module for this task is identical to the module used for the block stacking task, as shown in Figure~\ref{fig:language-module}.
The training and optimizer hyperparameters are the same as in the block stacking task.

As in the block stacking task, we tune the baseline representation hyperparameters over continuous sizes 20, 50, 100, 300, and 600, as well as an identical-sized discrete representation.

\end{document}